\documentclass[twoside,11pt]{article}

\usepackage{automl2019}
\usepackage{times}
\usepackage{epsfig}
\usepackage{graphicx}
\usepackage{wrapfig}
\usepackage{amsmath}
\usepackage{amssymb}
\usepackage[normalem]{ulem}
\usepackage{amsmath,}
\usepackage{algorithm}
\usepackage{algpseudocode}
\usepackage{pseudocode}
\usepackage{fancyhdr}
\usepackage{multirow}
\usepackage{caption}
\usepackage{hhline}
\usepackage{xcolor}
\usepackage{dblfloatfix}
\usepackage{subcaption}
\usepackage{enumitem}
\usepackage{booktabs}

\newcommand{\comm}[1]{}

\newcommand{\sys}{ASCAI} 

\def\delete{\comm}
\def\edit{\textcolor{black}}
\newcommand{\mli}[1]{\mathit{#1}}
\graphicspath{{figures/}} 

\firstpageno{1}

\begin{document}

\vspace{-1cm}
\title{
\vspace{-1cm}
\sys{}: Adaptive Sampling for acquiring Compact AI
\vspace{-0.5cm}}
\author{
Mojan Javaheripi\and Mohammad Samragh \and Tara Javidi \and Farinaz Koushanfar \\
\centering{
      \addr Department of Electrical and Computer Engineering, UC San Diego, USA} \\
      \centering{ \addr
      \{mojan, msamragh, tjavidi, farinaz\}@ucsd.edu} \\
    }
\maketitle

\thispagestyle{fancy}

\vspace{-0.6cm}
\begin{abstract}
\vspace{-0.6cm}

\noindent This paper introduces \sys{}, a novel \edit{adaptive sampling} methodology that can \textit{learn} how to effectively compress Deep Neural Networks (DNNs) for accelerated inference on resource-constrained platforms. Modern DNN compression techniques comprise various hyperparameters that require per-layer customization to ensure high accuracy. Choosing such hyperparameters is cumbersome as the pertinent search space grows exponentially with the number of model layers. To effectively traverse this large space, \edit{we devise an intelligent sampling mechanism that adapts the sampling strategy using customized operations inspired by genetic algorithms.} 
As a special case, we consider the space of model compression as a vector space. The adaptively selected samples enable
\sys{} to automatically \textit{learn} how to tune per-layer compression hyperparameters to optimize the accuracy/model-size trade-off. Our extensive evaluations show that \sys{} outperforms rule-based and reinforcement learning methods in terms of compression rate and/or accuracy. 

\end{abstract}

\setlength{\belowdisplayskip}{8pt} \setlength{\belowdisplayshortskip}{8pt}
\setlength{\abovedisplayskip}{8pt} \setlength{\abovedisplayshortskip}{8pt}

\vspace{-0.5cm}
\section{Introduction}\label{sec:intro}


\vspace{-0.2cm}
With the growing range of applications for Deep Neural Networks (DNNs) on embedded platforms, various DNN compression techniques have been developed to enable execution of such models on resource-limited devices. Some examples of DNN compression methods include pruning~[\cite{lin2018accelerating}], quantization~[\cite{zhou2016dorefa,ghasemzadeh2018rebnet}], nonlinear encoding~[\cite{han2015deep,samragh2019codex,samragh2017customizing}], and tensor decomposition~[\cite{kim2015compression,samragh2019autorank}].
\begin{wrapfigure}{r}{0.4\columnwidth}
    \centering
    \vspace{-0.2cm}
\includegraphics[width=0.39\columnwidth]{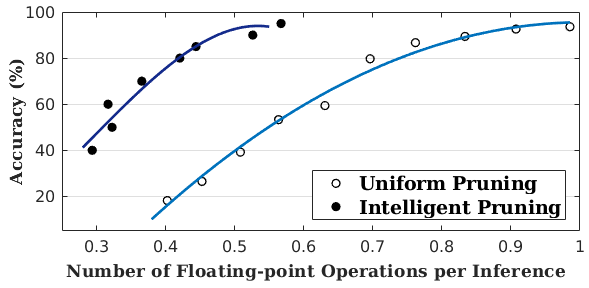}
    \vspace{-0.2cm}
    \caption{Accuracy/FLOPs Pareto frontiers for pruning a pre-trained VGG network on CIFAR-10. 
    Our intelligent policy achieves a better Pareto curve compared to rule-based (uniform) pruning.
    }
    \label{fig:pareto}
    \vspace{-0.3cm}
\end{wrapfigure}
Compression techniques require tailoring for each DNN architecture, which is generally realized by tuning certain hyperparameters at each layer.
Fig.~\ref{fig:pareto} illustrates the importance of intelligent hyperparameter selection for the example of pruning. 
In general, finding optimal hyperparameters is quite challenging as the space of possibilities grows exponentially with the number of DNN layers. Such large search-space renders manual or computerized greedy hyperparameter tuning algorithms sub-optimal or infeasible. 
Heuristics that compress one layer at a time overlook the existing inter-dependencies among layers. As such, an intelligent policy that globally tunes the pertinent hyperparameters for all layers is highly desired.

\edit{In this paper, we propose \sys{}, an adaptive sampling methodology that automates hyperparameter selection for DNN compression. Genetic algorithms are leveraged to iteratively adapt the sampling strategy.}
\delete{In this paper, we propose \sys{}, a method based on genetic algorithms (a.k.a., evolutionary strategies) that automates hyperparameter selection for DNN compression.} We devise a customized translator that encodes each compressed DNN as a fixed-length \edit{sample (vector) of the space}, called an \textit{individual}. \edit{We then adaptively sample from the pertinent vector space with the goal of finding individuals that render higher compression rate and inference accuracy.}
Our algorithm initializes a random \textit{population} of \delete{such} individuals and iteratively \textit{evolve}s them towards higher quality generations. To assess the quality of individuals, we develop a scoring mechanism that captures the trade-off between model accuracy and computational complexity.\delete{ The core idea behind \sys{} evolution is} \sys{} evolution aims to encourage survival of individuals with high scores and elimination of the weak ones. Towards this goal, each evolution iteration is modeled by a set of consecutive genetic operations. First, individuals with high scores are selected to generate a new population (\textit{evaluation} and \textit{selection}). Next, the chosen individuals are combined and perturbed to produce children of similar quality (\textit{crossover} and \textit{mutation}). The same procedure continues until convergence. 

\vspace{-0.5cm}
\section{Background and Related Work}\label{sec:related}
\vspace{-0.15cm}
\noindent{\bf Preliminaries and Key Insights.}
Recent research showcase the superiority of genetic algorithms to other exploration methods \edit{(e.g., reinforcement learning and random search)} for large search-spaces~[\cite{xie2017genetic}] due to their intriguing characteristics:
(1)~Genetic algorithms are inexpensive to implement as they do not rely on backpropagation. (2)~They are highly parallelizable. (3)~Genetic algorithms support a variety of scores and do not suffer in settings with sparse/discrete rewards~\edit{[\cite{such2017deep}]}. (4)~Finally, these methods can adopt multiple DNN compression tasks.
\begin{wrapfigure}{r}{0.45\columnwidth}
    \centering
    \vspace{-0.2cm}
    \includegraphics[width=0.43\columnwidth]{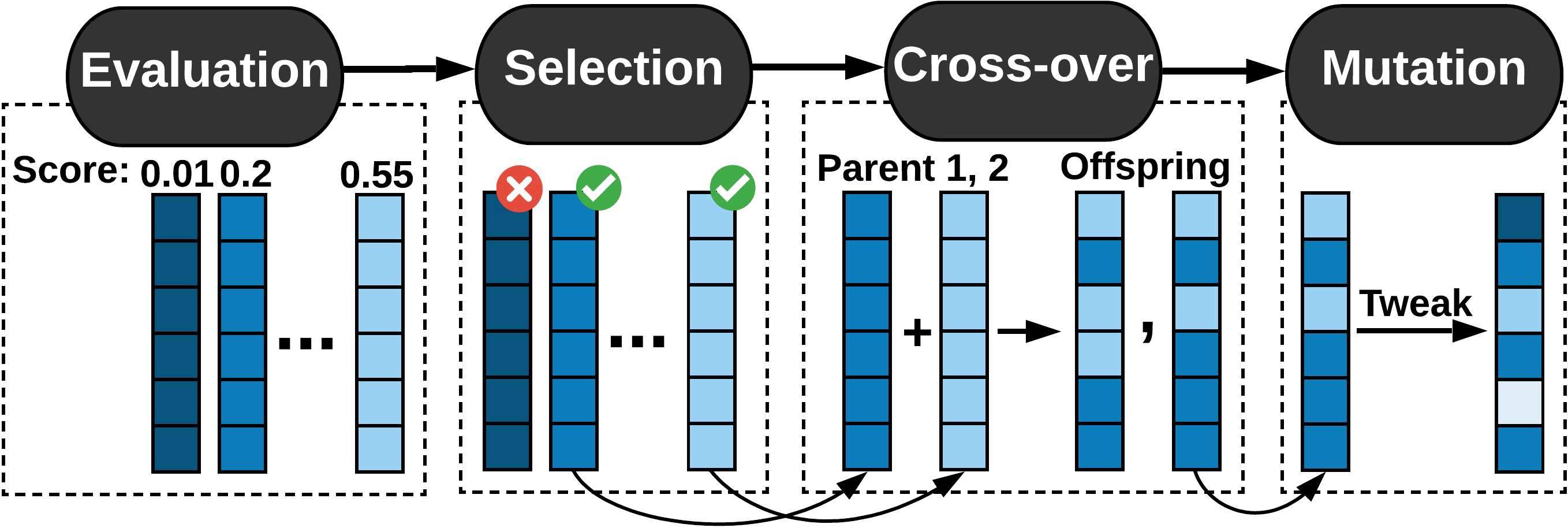}
    \vspace{-0.2cm}
    \caption{Operations of genetic algorithms.}
    \label{fig:genetic_ops}
    \vspace{-0.4cm}
\end{wrapfigure}
A genetic algorithm works on a \textit{generation} of \textit{individual}s.
The core idea is to encourage creation of superior individuals and elimination of the inferior ones. To this end, an iterative process (Fig.~\ref{fig:genetic_ops}) evolves the previous generation into a new, more competent population by performing a set of bio-inspired actions, i.e., \textit{selection}, \textit{crossover}, and \textit{mutation}.
During evaluation, individuals are assigned scores representing their quality (\textit{fitness}). The selection step performs a sampling (with replacement) based on individuals' fitness scores. Crossover and mutation create new individuals from existing ones.



\noindent{\bf{Related Work.}} In the context of deep learning, genetic algorithms have been applied to Neural Architecture Search (NAS)~[\cite{xie2017genetic,real2017large,huang2018data}], where the goal is to build a neural network architecture. The objective of NAS is generally achieving higher inference accuracy with no or little emphasis on the execution cost. Different from NAS, this paper focuses on learning hyperparameters for DNN customization, which simultaneously targets execution cost and inference accuracy.
Network compression has been studied in contemporary research~[\cite{he2018soft,he2017channel,wang2017structured,jiang2018efficient,li2016pruning,lin2017runtime,lin2018accelerating,luo2017thinet}]. 
These algorithms aim at eliminating redundancies from pre-trained network architectures to reduce computational complexity while preserving inference accuracy.
Reinforcement Learning (RL) is proposed as an automated tool that searches for improved model compression quality~[\cite{he2018amc}]. Although effective in finding near-optimal solutions, RL relies on gradient-based training, which can lead to a high computational burden and a slow convergence. 
[\cite{hu2018novel}] develop a novel pruning scheme that selects the pruned filters using genetic algorithms rather than magnitude-based or gradient-based approaches. Our work is different in that we aim to learn compression hyperparameters rather than the compression technique itself. As a result, \sys{} can be applied to generalized compression techniques.

\vspace{-0.5cm}
\section{\sys{} Approach}\label{sec:approach}
\begin{figure}[H]
    \centering
    \vspace{-0.2cm}
    \includegraphics[width=0.5\columnwidth]{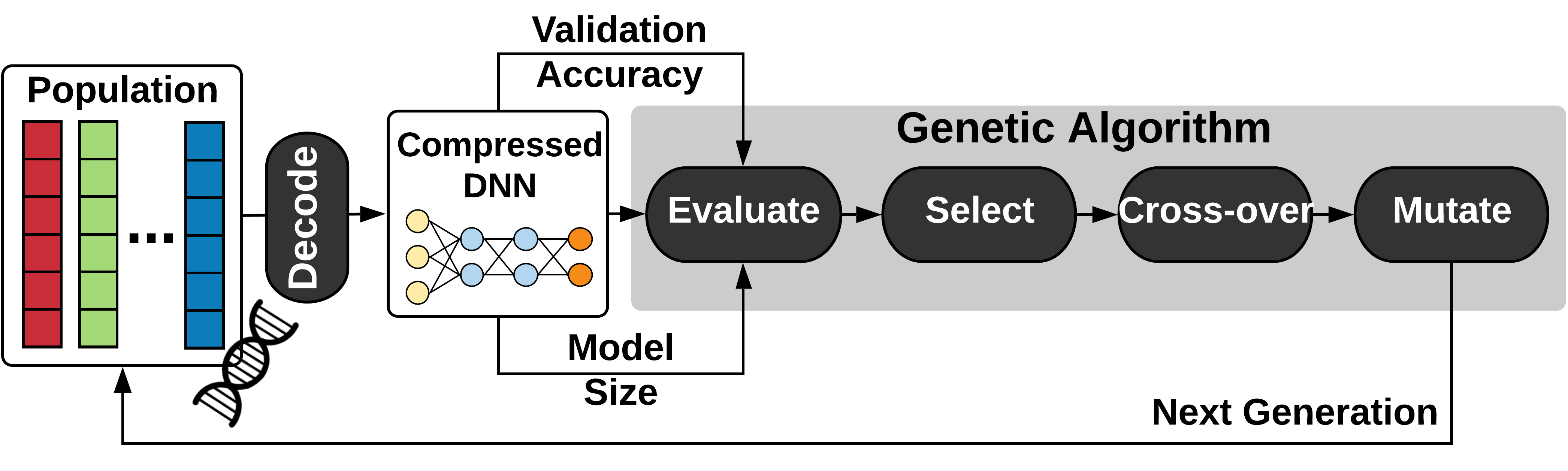}
    \vspace{-0.2cm}
    \caption{\sys{} solution for DNN customization.}
    \label{fig:overall}
\end{figure}

\vspace{-0.2cm}
An overview of \sys{} methodology is shown in Fig.~\ref{fig:overall}. A translation scheme represents each customized DNN using a vector of per-layer hyperparameters (Sec.~\ref{sec:translation}).
The initial population is established using a random sampling scheme (Sec.~\ref{sec:init}). At each iteration, the evaluation, selection, crossover, and mutation operations (Sec.~\ref{sec:gen_ops}) are performed to update the population towards a new generation.
By iteratively applying these operations, \sys{} finds near-optimal compression hyperparameters for a desired DNN.
\edit{The rest of this section provides details on the search algorithm}.


\vspace{-0.3cm}
\subsection{Genetic Translation}\label{sec:translation}


Although our proposed approach is applicable to various DNN compression techniques, in this paper we direct our focus on four compression tasks: structured and non-structured Pruning~[\cite{li2016pruning,he2017channel}], Singular Value Decomposition~[\cite{zhou2016less}], and Tucker-2 approximation~[\cite{kim2015compression}]. To construct the individual vectors for our genetic algorithm, we append per-layer hyperparameters as described in the following.



\noindent{\bf Pruning.}  We allocate a continuous value $p\in[0, 1]$ for each layer to represent the ratio of pruned values. For an $L$-layer DNN, each individual would be a vector $v\in \mathbb{R}^L$ with elements $v_i\in[0,1]$.

\noindent{\bf SVD.} We apply SVD on weight parameters ($W$) of fully-connected layers ($W \in \mathbb{R}^{m\times n}$) and point-wise convolutions ($W \in \mathbb{R}^{m\times n \times 1 \times 1}$). To represent the approximation rank in each layer, we discretize the possibilities into $64$ values and encode them as follows:
\begin{equation*}
        rank \in \{1\gets\frac{R}{64},\ 2\gets\frac{2R}{64},\ \dots,\ 64\gets R\},\ \  R = min\{m,n\}
\end{equation*}


\noindent{\bf Tucker-2.} Tucker decomposition is a generalized form of low-rank approximation for arbitrary-shaped tensors. We apply this method to 4-way weights in convolutional layers, $W\in \mathbb{R}^{m\times n \times k \times k}$. We focus on Tucker-2 which only decomposes the tensor along $m$ and $n$ directions, i.e., output and input channels. For each layer, a tuple of approximation ranks $(rank_m, rank_n)$ should be provided.
 We quantize the space of decomposition ranks to $8$ bins per-way as follows:
\begin{equation*}
    \begin{aligned}
        rank_m\in & \{1\gets\frac{m}{8},\ 2\gets\frac{2m}{8},\ \dots,\ 8\gets m\} ,\ \ \ \ \ 
        rank_n\in & \{1\gets\frac{n}{8},\ 2\gets\frac{2n}{8},\ \dots,\ 8\gets n\}
    \end{aligned}
\end{equation*}
When applying low-rank approximation, for a DNN that has a total of $L_1$ fully connected and point-wise convolutions ($1\times 1$ filters) and $L_2$ regular convolution layers ($k\times k$ filters), the individual would be a vector of length $L_1+2L_2$ that represent the encoded ranks described above.
\vspace{-0.3cm}
\subsection{Warm Initialization}\label{sec:init}
\vspace{-0.1cm}
A na\"ive initialization of individuals can result in a slow and sub-optimal convergence. To address this, we utilize \textit{warm} population initialization to reduce search time by eliminating unnecessary exploration of low-score regions, \edit{i.e., regions at which the inference accuracy is low.} 
Let us denote an individual vector as $v\in \mathbb{R}^{L}$, the validation dataset as $D=\{(x_m,y_m)\}_{m=1}^M$, and the corresponding classification accuracy as $acc_{D|v}$. 
During initialization, we only accept randomly sampled (\textit{i.i.d.}) individuals that satisfy an accuracy threshold: $acc_{D|v}>acc_{thr}$. To this end, we find a threshold vector $\theta$ that specifies the maximum per-layer compression when all other layers are uncompressed. Below we explain how to obtain $\theta$ for continuous and discrete hyperparameters.

\noindent{\bf Pruning.} We obtain a threshold vector $\theta \in \mathbb{R}^{L}$ with the $i$-th element $\theta_i$ specifying the maximum pruning rate for the $i$-th layer such that the accuracy threshold $acc_{thr}$ is not violated:
\begin{equation}\label{eq:prune_th_init}
\vspace{-0.2cm}
\theta_i = max\{p\}\ \ s.t. \ \ v_{j}=\begin{cases} 
      p & j = i \\
      0 & j \neq i 
   \end{cases}\  \&\ \  acc_{D|v}>acc_{thr}
\end{equation}
For each individual $v$, the $i$-th element $v_i$ is sampled from a Normal distribution $\mathcal{N}(\theta_i/2,\theta_i/2)$.

\noindent{\bf Decomposition.} The threshold vector $\theta$ represents per-layer minimum ranks that satisfy $acc_{thr}$:
\begin{equation}
\vspace{-0.2cm}
\theta_i = min\{rank\}\ \ s.t. \ \ v_{j}=\begin{cases} 
      rank & j = i \\
      rank_{max} & j \neq i 
   \end{cases}\  \&\ \  acc_{D|v}>acc_{thr}
\end{equation}
where $rank_{max}$ corresponds to the non-decomposed layer parameters (see Sec.~\ref{sec:translation}). Once $\theta$ is obtained, we uniformly sample $v_i$ from integers $\{\theta_i,\ \theta_i+1,\ \dots\}$.
\subsection{Genetic Operations}\label{sec:gen_ops}
To enable efficient exploration of the underlying search-space, we devise customized genetic operations, namely, evaluation, selection, crossover, and mutation. Below we describe each step.

\noindent{\bf Evaluation.} To assess a population of individuals, we first decode them to their corresponding compressed DNN architectures; this is done by applying compression to each layer of the original DNN based on the corresponding hyperparameter in the individual.
We develop a customized scoring mechanism that reflects the acquired model's accuracy and compression rate.
Given validation dataset, $D=\{(x_m,y_m)\}_{m=1}^M$, the score of individual $v$ evaluated on dataset $D$ is:
\begin{equation}\label{eq:score}
    score(v,D) = \frac{\Delta \mli{FLOPS}(v)}{\mli{PEN}_{acc}(D|v)}
\end{equation}
The numerator encourages reduction in model FLOPs while the denominator penalizes the decrease in model accuracy caused by compression. Here, $\Delta \mli{FLOPS}(v)$ represents the difference in FLOPS between the uncompressed DNN and the compressed model after applying $v$. $\mli{PEN}_{acc}(D|v)$ is a measure for accuracy degradation: 
\begin{equation}\label{eq:acc_penalty}
         \mli{PEN}_{acc}(D|v)=~acc_{D|o} - acc_{D|v}
         +~T(acc_{D|v}<acc_{thr})\times e^{acc_{thr} - acc_{D|v}}
\end{equation}
where $acc_{D|o}$ is the validation accuracy of the original (uncompressed) model, $acc_{D|v}$ is the accuracy after applying compression with $v$, and $T(.)$ returns the Boolean assessment of the provided inequality. To prevent undesirable drop of accuracy, we greatly diminish the score of individuals that cause lower accuracies than the set constraint, $acc_{thr}$; 
Having an accuracy constraint is crucial since the genetic algorithm will converge to a model size of zero otherwise. 
To ensure efficiency, we only use a small portion of the training samples as validation data.





\noindent{\bf Selection.} The selection stage in genetic algorithms attempts to choose high-quality individuals to generate the next population. Let us denote the population at the beginning of $t$-th iteration by $\mathbb P_{t}=\{v_{n}^t\}_{n=1}^N$ with fitness scores $\{s_n\}_{n=1}^N$ obtained from Eq.~\ref{eq:score}. Following~[\cite{xie2017genetic}], we normalize the scores as $s_n \gets \frac{s_n-s_{min}}{\sum_{n=1}^{N} (s_n-s_{min})}$, where $s_{min}$ is the minimum score in current population. Subtraction of the minimum score ensures that the probability of selecting the weakest individual is zero.
The new population, $\mathbb P_{t+1}=\{v_{n}^{t+1}\}_{n=1}^N$, is generated by non-uniform random sampling (with replacement) from the old population, $\mathbb P_t$. In this non-uniform sampling, the probability of selecting an individual, $v_n^t$, is proportional to its score. 
This method eliminates weak individuals and passes the high-quality ones to the next generation.

\noindent{\bf Crossover.} Given a selected population $\mathbb{P}_{t+1}$, crossover generates two offsprings by operating on each pair of adjacent parent individuals $\{v_{2k-1}^{t+1},v_{2k}^{t+1}\}_{k=1}^{\frac{N}{2}}$. We use two parameters to control the degree of crossover operation: $p_{cross}$ determines the probability of applying crossover between two individuals, and $p_{swap}$ is the per-element swapping probability. Crossover allows superior individuals to exchange their learned patterns and enables knowledge transfer across the population.

\noindent{\bf Mutation.} 
Mutation randomly tweaks each individual in the population. Similar to crossover, we define two control parameters: $p_{mutate}$ is the probability that the individual gets mutated and $p_{tweak}$ determines the per-element tweaking probability. Mutation allows exploration of the neighborhood of candidate points in the search-space. Each element of a continuous-valued individual is mutated by adding a random value drawn from a zero-mean Normal distribution $\mathcal{N}(0,\,0.2)$. Discrete-valued individuals are mutated by randomly incrementing or decrementing vector elements. The values are clipped to the valid ranges after mutation.


\vspace{-0.2cm}
\section{Experiments}\label{sec:experiments}


We provide extensive evaluations on CIFAR-10 and ImageNet benchmarks. 
The baseline networks are trained from scratch using PyTorch library.
We conduct experiments with non-structured pruning ($P_n$), structured pruning ($P_s$), decomposition ($D$), and combination of multiple compression methods ($D+P_s$).
 Tab.~\ref{tab:comparison-results} summarizes the results of \sys{} compressed networks and compares them with prior work that utilize pruning as the compression technique. For brevity, we compare \sys{} with best existing works and exclude other related works.

\begin{table}[h]
\vspace{-0.2cm}
\centering
\caption{Comparison of \sys{} with state-of-the-art compression methods, namely CP~[\cite{he2017channel}], AMC~[\cite{he2018amc}], SFP~[\cite{he2018soft}], FP~[\cite{li2016pruning}], SSS~[\cite{wang2017structured}], GDP~[\cite{lin2018accelerating}], ThiNet~[\cite{luo2017thinet}], and RNP~[\cite{lin2017runtime}]. For ImageNet, we follow common practice in prior work and compare our top-5 accuracy with them. }\label{tab:comparison-results}
\vspace{-0.9cm}
\end{table}
\begin{figure}[h]
    \centering
    \vspace{-0.3cm}
\includegraphics[width=1\columnwidth]{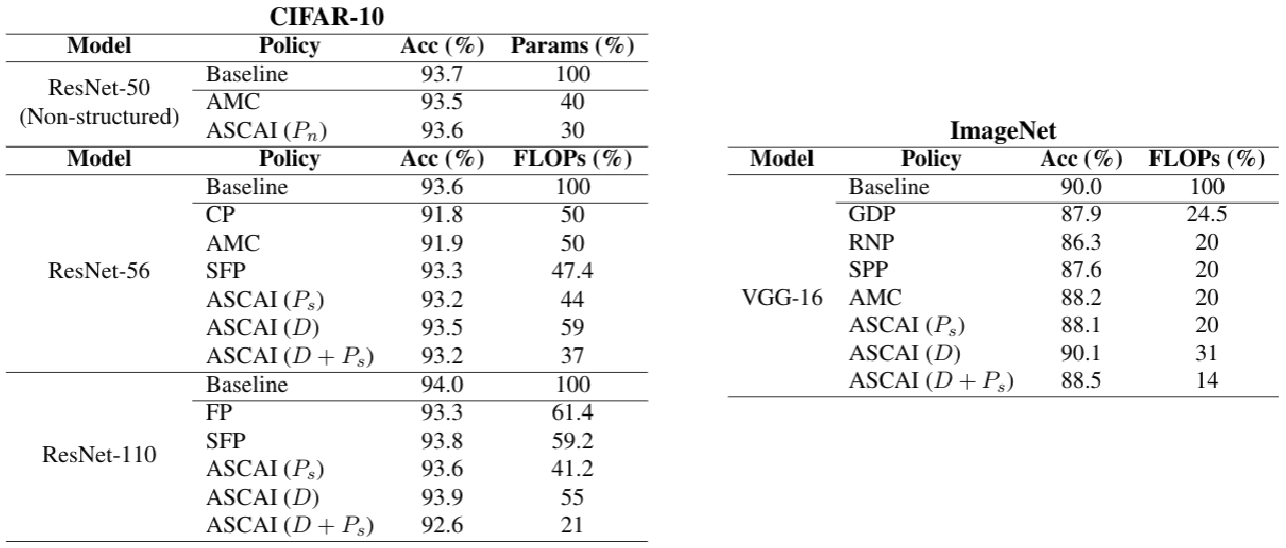}
    \vspace{-0.5cm}
    \vspace{-0.1cm}
\end{figure}

\noindent\textbf{Non-structured Pruning ($P_n$ in Tab.~\ref{tab:comparison-results}).} We perform non-structured pruning on ResNet-50 trained on CIFAR-10. We prune the parameters with lowest absolute value as in AMC~[\cite{han2015deep}], the state-of-the-art that utilizes RL for automated DNN compression. Similar to AMC, we do not perform fine-tuning on the compressed model in this experiment. As shown, \sys{} achieves better accuracy with $1.33\times$ lower parameters.

\noindent\textbf{Structured Pruning ($P_s$ in Tab.~\ref{tab:comparison-results}).} We implement structured pruning by adding masks after $ReLU$ activation layers. Following~[\cite{molchanov2016pruning}], we prune the activations in each layer based on the sum of absolute gradients at the ReLU output.
We base our comparisons on the number of operations per inference, i.e., FLOPs, compared to the uncompressed baseline. On CIFAR-10 networks, \sys{} achieves on average $1.25\times$ lower FLOPs, while achieving similar classification accuracy compared to prior art. \delete{On ResNet-50 trained for ImageNet, \sys{} compresses the model to $2.1\%$ less FLOPs while achieving $1.4\%$ higher top-5 accuracy compared to the best prior work~[\cite{lin2018accelerating}].}
On VGG-16, \sys{} outperforms all heuristic methods and gives competing results with AMC~[\cite{he2018amc}]. 

\noindent\textbf{Decomposition and Pruning ($D+P_s$ in Tab.~\ref{tab:comparison-results}).} To unveil the full potential of our method, we allow \sys{} to learn and combine multiple compression techniques, namely, structures pruning, SVD, and Tucker decomposition.  We also report the FLOPs reduction achieved by decomposition separately (shown by $D$ in Tab.~\ref{tab:comparison-results}). Combining multiple techniques allows \sys{} to push the limits of compression. \delete{For instance, on ResNet-50 trained for ImageNet, \sys{} achieves an average of $2\times$ lower FLOPs than related work while having a slightly higher accuracy.} On VGG-16, \sys{} pushes the state-of-the-art FLOPs reduction from $5\times$ to $7.2\times$ with $0.3\%$ higher accuracy.

\vspace{-0.5cm}
\subsection{Analysis and Discussion}\label{sec:discussion}
\begin{figure}[h]
    \centering
    \vspace{-0.3cm}
\includegraphics[width=1\columnwidth]{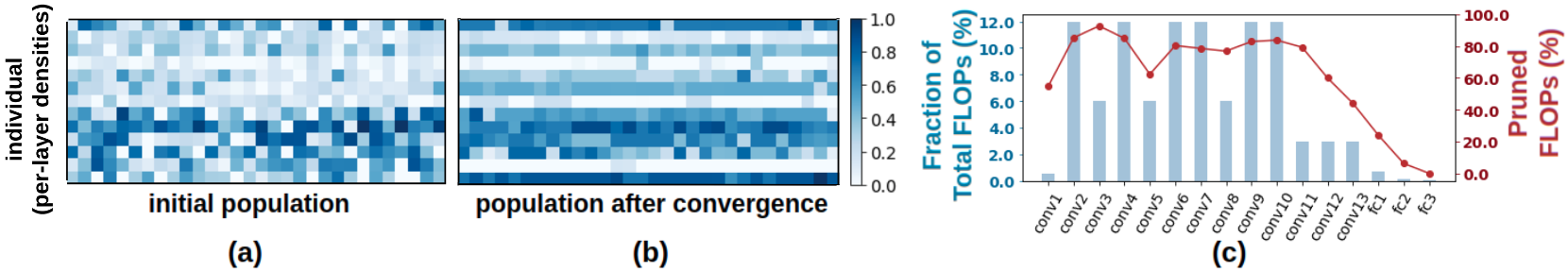}
    \vspace{-0.5cm}
    \caption{(a)~Initialized population for structured pruning. (b)~Population upon convergence. Here, each row corresponds to a DNN layer and each column denotes an individual in the population. (c)~Per-layer FLOPs (bar charts) and percentage of pruned FLOPs (curve).}
    \label{fig:discussion}
    \vspace{-0.5cm}
\end{figure}

To illustrate \sys{} methodology, we consider VGG architecture trained on ImageNet and compressed with structured (filter) pruning. The population is visualized at the initial step (Fig.~\ref{fig:discussion}-a) and after 50 iterations of genetic updating (Fig~\ref{fig:discussion}-b). Upon convergence, individuals strongly resemble one another and have similarly high scores. 
\sys{} successfully learns expert-designed rules: first and last rows in Fig.~\ref{fig:discussion}-b (first convolution and last fully-connected) are given high densities to maintain inference accuracy.
\sys{} performs \textit{whole-network} compression by capturing the state of all layers in each genetic individual. As such, our algorithm can learn which configuration of hyperparameters least affects model accuracy and most reduces the overall FLOPs. To show this capability, we present the per-layer FLOPs for VGG-16 network trained on ImageNet in Fig.~\ref{fig:discussion}-c. The bar for each layer shows the percentage of total FLOPs in the original model; the curve shows the percentage of pruned FLOPs in the compressed network. 
Different from prior art~[\cite{jiang2018efficient}], \sys{} prunes the first convolutions more and relaxes pruning for late convolution and fully-connected layers as they have a minor role in FLOPs. 



\begin{wrapfigure}{r}{0.35\columnwidth}
    \centering
    \vspace{-1cm}
\includegraphics[width=0.33\textwidth]{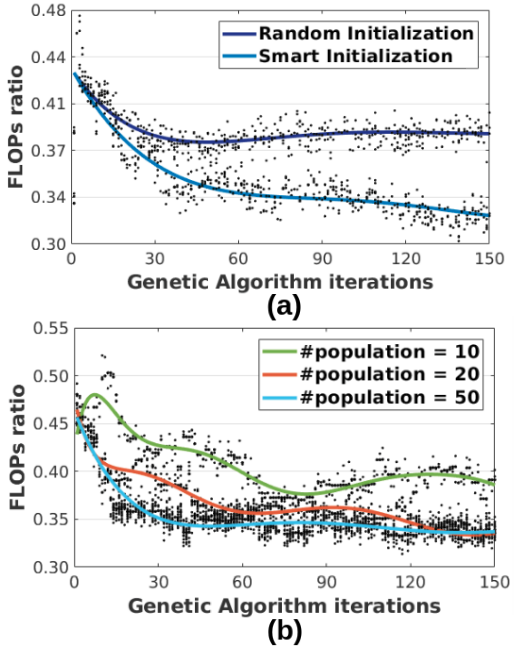}
    \vspace{-0.3cm}
    \caption{(a) Effect of initialization. (b) Effect of population size. 
    }
    \label{fig:ablation}
    \vspace{-1cm}
\end{wrapfigure}


\vspace{-0.3cm}
\subsection{Ablation Study}\label{sec:ablation}
\vspace{-0.1cm}
In this section, we study the effect of \sys{} components on algorithm convergence and final FLOPs/accuracy. For brevity, we only focus on structured pruning for CIFAR-10. We show the trend lines as well as a fraction of individuals (black dots) across \sys{} iterations.

\noindent\textbf{Effect of Initialization.} 
Fig.~\ref{fig:ablation}-a shows the evolution of FLOPs ratio for two initialization policies, one with uniformly random samples and one with our proposed initialization scheme discussed in Sec.~\ref{sec:init}. As seen, naive initialization greatly harms the convergence rate and final FLOPs.


\noindent\textbf{Effect of Population Size.}  Fig.~\ref{fig:ablation}-b presents the effect of population size on \sys{} convergence. A higher number of individuals results in a smoother convergence and lower final FLOPs. This effect saturates for a large enough population. 
\vspace{-0.3cm}
\section{Conclusion}\label{sec:conclusion}
This paper introduces \sys{}, a method to automate DNN compression using adaptive sampling with genetic algorithms. Our algorithm learns how the compression hyperparameters should be set across layers to achieve a better performance than models designed by human experts. The core idea behind \sys{} is to translate compression hyperparameters into a vector of genes and explore the corresponding search-space using genetic operations.
This approach allows \sys{} to be generic and applicable to any combination of post-processing DNN compression methods. 


\vskip 0.2in
\bibliography{sample}

\begin{thebibliography}{23}
\providecommand{\natexlab}[1]{#1}
\providecommand{\url}[1]{\texttt{#1}}
\expandafter\ifx\csname urlstyle\endcsname\relax
  \providecommand{\doi}[1]{doi: #1}\else
  \providecommand{\doi}{doi: \begingroup \urlstyle{rm}\Url}\fi

\bibitem[Ghasemzadeh et~al.(2018)Ghasemzadeh, Samragh, and
  Koushanfar]{ghasemzadeh2018rebnet}
Mohammad Ghasemzadeh, Mohammad Samragh, and Farinaz Koushanfar.
\newblock Rebnet: Residual binarized neural network.
\newblock In \emph{2018 IEEE 26th Annual International Symposium on
  Field-Programmable Custom Computing Machines (FCCM)}, pages 57--64. IEEE,
  2018.

\bibitem[Han et~al.(2015)Han, Mao, and Dally]{han2015deep}
Song Han, Huizi Mao, and William~J Dally.
\newblock Deep compression: Compressing deep neural networks with pruning,
  trained quantization and huffman coding.
\newblock \emph{arXiv preprint arXiv:1510.00149}, 2015.

\bibitem[He et~al.(2018{\natexlab{a}})He, Kang, Dong, Fu, and Yang]{he2018soft}
Yang He, Guoliang Kang, Xuanyi Dong, Yanwei Fu, and Yi~Yang.
\newblock Soft filter pruning for accelerating deep convolutional neural
  networks.
\newblock \emph{arXiv preprint arXiv:1808.06866}, 2018{\natexlab{a}}.

\bibitem[He et~al.(2017)He, Zhang, and Sun]{he2017channel}
Yihui He, Xiangyu Zhang, and Jian Sun.
\newblock Channel pruning for accelerating very deep neural networks.
\newblock In \emph{International Conference on Computer Vision (ICCV)},
  volume~2, 2017.

\bibitem[He et~al.(2018{\natexlab{b}})He, Lin, Liu, Wang, Li, and
  Han]{he2018amc}
Yihui He, Ji~Lin, Zhijian Liu, Hanrui Wang, Li-Jia Li, and Song Han.
\newblock Amc: Automl for model compression and acceleration on mobile devices.
\newblock In \emph{Proceedings of the European Conference on Computer Vision
  (ECCV)}, pages 784--800, 2018{\natexlab{b}}.

\bibitem[Hu et~al.(2018)Hu, Sun, Li, Wang, and Gu]{hu2018novel}
Yiming Hu, Siyang Sun, Jianquan Li, Xingang Wang, and Qingyi Gu.
\newblock A novel channel pruning method for deep neural network compression.
\newblock \emph{arXiv preprint arXiv:1805.11394}, 2018.

\bibitem[Huang and Wang(2018)]{huang2018data}
Zehao Huang and Naiyan Wang.
\newblock Data-driven sparse structure selection for deep neural networks.
\newblock In \emph{Proceedings of the European Conference on Computer Vision
  (ECCV)}, pages 304--320, 2018.

\bibitem[Jiang et~al.(2018)Jiang, Li, Qian, and Tang]{jiang2018efficient}
Chunhui Jiang, Guiying Li, Chao Qian, and Ke~Tang.
\newblock Efficient dnn neuron pruning by minimizing layer-wise nonlinear
  reconstruction error.
\newblock In \emph{IJCAI}, pages 2--2, 2018.

\bibitem[Kim et~al.(2015)Kim, Park, Yoo, Choi, Yang, and
  Shin]{kim2015compression}
Yong-Deok Kim, Eunhyeok Park, Sungjoo Yoo, Taelim Choi, Lu~Yang, and Dongjun
  Shin.
\newblock Compression of deep convolutional neural networks for fast and low
  power mobile applications.
\newblock \emph{arXiv preprint arXiv:1511.06530}, 2015.

\bibitem[Li et~al.(2016)Li, Kadav, Durdanovic, Samet, and Graf]{li2016pruning}
Hao Li, Asim Kadav, Igor Durdanovic, Hanan Samet, and Hans~Peter Graf.
\newblock Pruning filters for efficient convnets.
\newblock \emph{arXiv preprint arXiv:1608.08710}, 2016.

\bibitem[Lin et~al.(2017)Lin, Rao, Lu, and Zhou]{lin2017runtime}
Ji~Lin, Yongming Rao, Jiwen Lu, and Jie Zhou.
\newblock Runtime neural pruning.
\newblock In \emph{Advances in Neural Information Processing Systems}, pages
  2181--2191, 2017.

\bibitem[Lin et~al.(2018)Lin, Ji, Li, Wu, Huang, and
  Zhang]{lin2018accelerating}
Shaohui Lin, Rongrong Ji, Yuchao Li, Yongjian Wu, Feiyue Huang, and Baochang
  Zhang.
\newblock Accelerating convolutional networks via global \& dynamic filter
  pruning.
\newblock In \emph{IJCAI}, pages 2425--2432, 2018.

\bibitem[Luo et~al.(2017)Luo, Wu, and Lin]{luo2017thinet}
Jian-Hao Luo, Jianxin Wu, and Weiyao Lin.
\newblock Thinet: A filter level pruning method for deep neural network
  compression.
\newblock In \emph{Proceedings of the IEEE international conference on computer
  vision}, pages 5058--5066, 2017.

\bibitem[Molchanov et~al.(2016)Molchanov, Tyree, Karras, Aila, and
  Kautz]{molchanov2016pruning}
Pavlo Molchanov, Stephen Tyree, Tero Karras, Timo Aila, and Jan Kautz.
\newblock Pruning convolutional neural networks for resource efficient transfer
  learning.
\newblock \emph{arXiv preprint arXiv:1611.06440}, 3, 2016.

\bibitem[Real et~al.(2017)Real, Moore, Selle, Saxena, Suematsu, Tan, Le, and
  Kurakin]{real2017large}
Esteban Real, Sherry Moore, Andrew Selle, Saurabh Saxena, Yutaka~Leon Suematsu,
  Jie Tan, Quoc~V Le, and Alexey Kurakin.
\newblock Large-scale evolution of image classifiers.
\newblock In \emph{Proceedings of the 34th International Conference on Machine
  Learning-Volume 70}, pages 2902--2911. JMLR. org, 2017.

\bibitem[Samragh et~al.(2017)Samragh, Ghasemzadeh, and
  Koushanfar]{samragh2017customizing}
Mohammad Samragh, Mohammad Ghasemzadeh, and Farinaz Koushanfar.
\newblock Customizing neural networks for efficient fpga implementation.
\newblock In \emph{Field-Programmable Custom Computing Machines (FCCM), 2017
  IEEE 25th Annual International Symposium on}, pages 85--92. IEEE, 2017.

\bibitem[Samragh et~al.(2019{\natexlab{a}})Samragh, Javaheripi, and
  Koushanfar]{samragh2019autorank}
Mohammad Samragh, Mojan Javaheripi, and Farinaz Koushanfar.
\newblock Autorank: Automated rank selection for effective neural network
  customization.
\newblock \emph{ML-for-Systems workshop at the 46th International Symposium on
  Computer Architecture (ISCA)}, 2019{\natexlab{a}}.

\bibitem[Samragh et~al.(2019{\natexlab{b}})Samragh, Javaheripi, and
  Koushanfar]{samragh2019codex}
Mohammad Samragh, Mojan Javaheripi, and Farinaz Koushanfar.
\newblock Codex: Bit-flexible encoding for streaming-based fpga acceleration of
  dnns.
\newblock \emph{arXiv preprint arXiv:1901.05582}, 2019{\natexlab{b}}.

\bibitem[Such et~al.(2017)Such, Madhavan, Conti, Lehman, Stanley, and
  Clune]{such2017deep}
Felipe~Petroski Such, Vashisht Madhavan, Edoardo Conti, Joel Lehman, Kenneth~O
  Stanley, and Jeff Clune.
\newblock Deep neuroevolution: Genetic algorithms are a competitive alternative
  for training deep neural networks for reinforcement learning.
\newblock \emph{arXiv preprint arXiv:1712.06567}, 2017.

\bibitem[Wang et~al.(2017)Wang, Zhang, Wang, and Hu]{wang2017structured}
Huan Wang, Qiming Zhang, Yuehai Wang, and Haoji Hu.
\newblock Structured probabilistic pruning for convolutional neural network
  acceleration.
\newblock \emph{arXiv preprint arXiv:1709.06994}, 2017.

\bibitem[Xie and Yuille(2017)]{xie2017genetic}
Lingxi Xie and Alan Yuille.
\newblock Genetic cnn.
\newblock \emph{arXiv preprint arXiv:1703.01513}, 2017.

\bibitem[Zhou et~al.(2016{\natexlab{a}})Zhou, Alvarez, and
  Porikli]{zhou2016less}
Hao Zhou, Jose~M Alvarez, and Fatih Porikli.
\newblock Less is more: Towards compact cnns.
\newblock In \emph{European Conference on Computer Vision}, pages 662--677.
  Springer, 2016{\natexlab{a}}.

\bibitem[Zhou et~al.(2016{\natexlab{b}})Zhou, Wu, Ni, Zhou, Wen, and
  Zou]{zhou2016dorefa}
Shuchang Zhou, Yuxin Wu, Zekun Ni, Xinyu Zhou, He~Wen, and Yuheng Zou.
\newblock Dorefa-net: Training low bitwidth convolutional neural networks with
  low bitwidth gradients.
\newblock \emph{arXiv preprint arXiv:1606.06160}, 2016{\natexlab{b}}.

\end{thebibliography}






\end{document}